# GreenEye: Development of Real-Time Traffic Signal Recognition System for Visual Impairments

Danu Kim

Korea International School, Jeju Campus, Jeju-do 63644, Republic of Korea
dukim27@kis.ac

## Abstract

Recognizing a traffic signal, determining if the signal is green or red, and figuring out the time left to cross the crosswalk are significant challenges to visually impaired people. Previous research has focused on recognizing only two traffic signals, green and red lights, using machine learning techniques. The proposed method developed a GreenEye system that recognizes the traffic signals' color and tells the time left for pedestrians to cross the crosswalk in real-time. GreenEye's first training showed the highest precision of 74.6%; four classes reported 40% or lower recognition precision in this training session. The data imbalance caused low precision; thus, extra labeling and database formation were performed to stabilize the number of images between different classes. After the stabilization, all 14 classes showed excelling precision rate of 99.5%.

## 1. Introduction

There are about 117,000 crosswalks in Korea (2021, [1]); most non-disabled people don't have challenges recognizing the traffic signal and walking the crosswalk. However, in cases of visually impaired people, recognizing the traffic signal and determining whether the traffic light is green or red to cross the crosswalk safely is a big challenge [1]. Further, knowing the time left to cross the crosswalk is a more significant challenge for visually impaired people. Across the traffic lights in the Korean peninsula, only 34% have beacons for visually impaired people. Seoul showed a 66.1% beacon installation rate, whereas Ulsan showed a 7.8% immensely lower installation rate. Moreover, the number of broken or not-working traffic lights has been 4,451 for four years until 2021, and it even took 184 days maximum to fix the traffic light [1, 2, 3].

As one of the representative automated traffic light recognition systems, Handong University's previous research [4] developed an algorithm that used machine learning techniques to recognize green and red traffic lights. However, the algorithm didn't have a feature for telling the time left to cross the crosswalk, which leaves a limitation for visually impaired people when used in real life. Further, other research gave relatively lower precision and training results in recognition. In contrast, the presented study used a further-developed YOLOv5 to derive increased precision and faster learning speed.

Therefore, the presented study proposed a real-time traffic light number auto-recognition system, GreenEye, to help visually impaired people safely and conveniently cross the crosswalk. The proposed method gives the time left to cross the crosswalk in an actual situation, in addition to the green and red light determination, which helps visually impaired people cross the road safer. A large amount of traffic light data was taken to develop the system on a real road, and those data were labeled later. Then, deep learning–based object detection was applied to recognize traffic lights from input traffic light images and videos in real-time, recognizing the colored lights, including the time left for the green light.

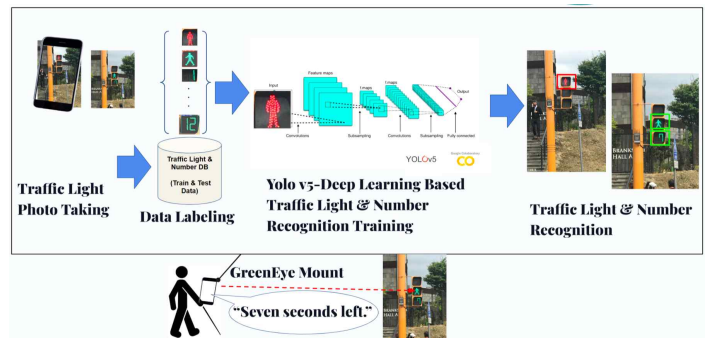

Figure 1. Proposed GreenEye System

## 2. Proposed GreenEye System

Fig. 1 is the diagram of the GreenEye System. First, a large amount of the traffic light data(Green, Red, and traffic signal number) were collected with the smartphone. Then, each traffic light image was labeled(custom data annotation) in the collected data. Using the custom data set with all the labeling processes completed as input, GreenEye was trained with YOLOv5 to recognize the color of the traffic light and the traffic signal number. Lastly, GreenEye was tested to recognize Green, Red, and traffic signal numbers on real-time videos about the traffic signals.

### 2.1 GreenEye System Component Processes
### 2.1.1 Custom Data Set Formation and Labeling

Conventional data related to traffic light recognition was labeled only on the whole traffic light or Green/Red lights, so the proposed research had to collect a vast amount of traffic light and traffic signal number images with a smartphone, forming a custom data set consisting of 1,113 images. Later, using LabelImg software [5], each object was labeled with a bounding box, coordinate, and class information.

### 2.1.2 Traffic Signal Number Training and Recognition

The proposed research employed YOLOv5 to achieve real-time object detection [6]. The GreenEye System is trained to categorize classes into green, red, and traffic signal numbers, and later, the system recognizes if any of those classes exist in new traffic light images or videos. Through the training and recognition in Fig. 1, the light is shown as 'Green Light,' 'Green Light_7' on the screen, performing a successful recognition.

GreenEye is based on YOLOv5 and is implemented in the Google Colab [7]. Colab has benefits, as it provides free cloud services for machine learning and deep learning implementation. Since large-scale training is operated on Google servers, the development process can be fast and easily tested without being limited to the computational resources of local computers.

## 3. Experimental Results

### 3.1 Experimental Setting

The custom data set, which consists of 884 images used in the first experiment, was collected at a nearby road site. This input data is used for training and validating the YOLOv5 traffic light recognition model. The proposed work analyzed various experimental results by setting the training and validation data ratio differently. Further, the dataset consists of 14 classes, including Green, Red, and signal numbers (1–12).

The final-trained YOLOv5 model had 214 layers and about 7 million parameters. For training efficiency, we set the input image size to 640, epoch to 30, and batch size to 16. The Colab GPU used in the experiment was the V100 Nvidia Premium GPU. Considering that GreenEye will be operated on a smartphone later, we set it as YOLOv5s, the small model, and it was trained to override the 80 original classes with 14 new classes. For deep learning model training, the training and validation ratio was divided into 7:3 and 9:1.

### 3.2 Test Results by Validation Ratio

Table 1 (left) shows the recognition precision of the GreenEye where the ratio of training to validation data is 7:3 and 9:1. The total duration for 30 epochs for 7:3 and 9:2 training was 2.2 and 2.4 hours each.

In 7:3 training and validation, the mAP50, the average precision for whole class recognition was 0.64. This precision shows that the model trained with 70% of input images correctly recognized 64% of Green light, Red light, and traffic signal numbers. However, the model had lower precision for traffic signal numbers Green_8, Green_9, and Green_12.

| Precision | Imbalanced Data | | Balanced Data | |
|---|---|---|---|---|
| | 7:3 | 9:1 | 7:3 | 9:1 |
| Green Light | 0.99 | 0.99 | 0.99 | 0.99 |
| Green_1 | 0.99 | 0.00 | 0.99 | 0.99 |
| Green_2 | 0.57 | 0.79 | 0.88 | 0.99 |
| Green_3 | 0.99 | 0.99 | 0.99 | 0.99 |
| Green_4 | 0.83 | 0.98 | 0.99 | 0.99 |
| Green_5 | 0.33 | 0.95 | 0.95 | 0.99 |
| Green_6 | 0.70 | 0.96 | 0.48 | 0.99 |
| Green_7 | 0.75 | 0.31 | 0.99 | 0.99 |
| Green_8 | 0.18 | 0.29 | 0.97 | 0.99 |
| Green_9 | 0.06 | 0.14 | 0.99 | 0.99 |
| Green_10 | 0.78 | 0.99 | 0.99 | 0.99 |
| Green_11 | 0.76 | 0.99 | 0.99 | 0.99 |
| Green_12 | 0.05 | 0.05 | 0.98 | 0.99 |
| Red Light | 0.99 | 0.99 | 0.99 | 0.99 |
| mAP | 0.64 | 0.75 | 0.94 | 0.99 |
| Train Time | 2.2 hr | 2.4 hr | 0.9 hr | 1.1 hr |

Table 1. Precision under Data Imbalance vs. Balance

In 9:1 setting, mAP50 increased from 7:3 validation to 75%. This increase shows that the model parameter training became more precise as more training data was used. Classes with successful recognition had increased precision, but classes with low precision from the previous experiment (Green_7, Green_8, Green_9, Green_12) still had low precision, which shows the necessity for improvement; Green_7 and Green_12 had about 30% and 5% recognition precision, respectively, offering a low precision of the GreenEye model.

### 3.3 Test Results after Resolving Data Imbalance

As mentioned earlier, the highest precision rate was 75% from 9:1 cross-validation; even though the training size was increased, the classes with low precision rates from 7:3 validation still showed low precision, which led to the in-depth examination of the data distribution, focusing on low-precision classes. Classes with high precision(e.g., Green_4) had about 80 input data. However, classes with low precision had a small number of data: about 50 input images. Green_9 class had 34 data, and Green_12 had 24 data.

These data imbalances could bring a decreased performance to deep learning training. Classes with a smaller number of data might have a smaller portion of training data than classes with more data, causing the labeled class not to be trained precisely. Therefore, to balance the overall data distribution between each class, extra photo taking and labeling for class data with low precision, including Green_7, proceeded. The total number of data increased from 884 to 1,113.

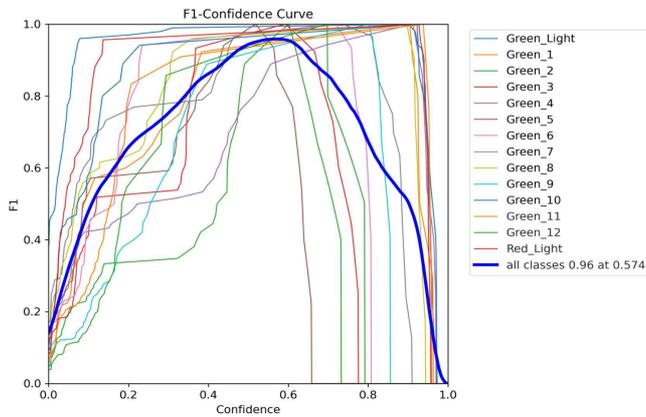

Figure 2. GreenEye System's F1 result with balanced data.

**Traffic Light Recognition in Videos [Training:Validation = 9:1]**

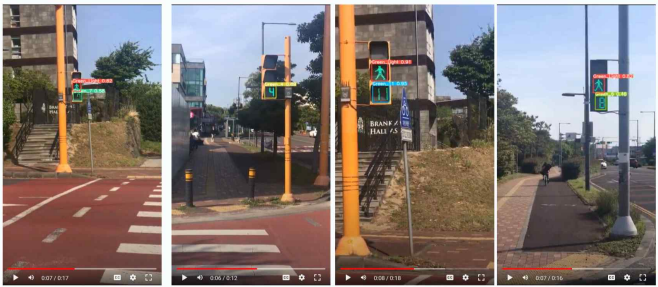

Figure 3. GreenEye performance result in real-life videos

Table 1 (Right) shows recognition results where data imbalance was resolved, and the training and validation image ratios were 7:3 and 9:1. 7:3 cross-validation's average recognition precision was 94%, showing a considerable increase from the previous experiments. Classes that showed low precision, such as Green_7, Green_8, and Green_9, showed a precision of 97%~99%. Training duration was 0.9 hours, showing that although input data increased, training time had decreased. This decrease in training duration results from a decrease in error during model parameter training for each class due to the stabilized data.

For 9:1 setting, the precision was 99%, with 1.1 hours of training duration, similar to the 7:3 test. All 14 classes recorded 99% of recognition precision. Fig. 2 shows the F1 precision of the proposed system. The F1 value is a performance criterion that checks the precision and recall values. With 57.4% confidence, the final F1 value recorded a high value of 96%.

### 3.4 Real-time GreenEye Tests on Traffic Light Videos

The GreenEye system should be used by visually impaired people on real-life roads, indicating that it should work without error on videos with real-time inputs. Fig. 3 shows the real-time recognition results of traffic light video on real roads containing Green, Red, and traffic signal number lights. The results show that it successfully recognized Green Light and traffic signal numbers. (video demo. https://youtu.be/BmQJiwo70n8)

| Precision | Previous [4] | GreenEye |
|---|---|---|
| Red Light | 97.1 % | 99.5 % |
| Green Light | 100.0 % | 99.5 % |
| Average | 98.52 % | 99.5 % |

Table 2. Precision comparison with previous research

| Performance | Imbalanced Data | Balanced Data |
|---|---|---|
| mAP@0.5 | 0.75 | 0.99 |
| Train Time | 2.4 hr | 1.1 hr |

Table 3. Performance improvement in GreenEye system

### 3.5 Performance Comparison with Previous Research

As shown in Table 2, the GreenEye system achieved a Green and Red light recognition precision of 99.5%, which is a better result than those from the previous research [4]. Table 3 shows that by tackling the data imbalance between the traffic light classes, GreenEye improved from prediction measured during the data imbalance of 75% precision to 99% precision for all 14 traffic light classes. Further, the training time was reduced from 2.39 hours to 1.13 hours.

### 4. Conclusion

The GreenEye is a real-time traffic signal recognition model that visually impaired people could use. The Google Colab and YOLOv5 model was customized to achieve this goal, recognizing all 14 traffic signal classes successfully. Further, it was observed that the system successfully works for the videos, too, showing the possibility of recognizing the traffic light well in portable input devices such as smartphone cameras. The future application of this research is to develop a smartphone app with a GreenEye system based on a TTS engine so that visually impaired people can use the traffic light recognition system more conveniently.